\documentclass{article}

\usepackage{PRIMEarxiv}

\usepackage[utf8]{inputenc} 
\usepackage[T1]{fontenc}    

\usepackage{url}            
\usepackage{booktabs}       
\usepackage{amsfonts}       
\usepackage{nicefrac}       
\usepackage{microtype}      
\usepackage{lipsum}
\usepackage{fancyhdr}       
\usepackage{graphicx}       
\graphicspath{{media/}}     

\usepackage{longtable}
\usepackage{soul}
\usepackage{xcolor}
\usepackage{float}
\usepackage{amsmath,amssymb,amsthm}
\usepackage{iftex}

\usepackage{graphicx}
\usepackage{float}

\usepackage{natbib}
\usepackage{hyperref}       
\title{RATSIFT: Resolution-Adaptive Transformer for Solar Irradiance Forecasting and Temporal Downscaling}

\author{
  Jesto Peter \\
Business Analytics, Economics, and Information System \\
  Bowling Green State University,\\
   Bowling Green, Ohio\\
    \texttt{jestop@bgsu.edu} \\
   \And
 Timothy M. Hansen \\
 Department of Electrical and Computer Engineering\\
      Walter Scott, Jr. College of Engineering\\
      Colorado State University, Fort Collins\\\\
      \texttt{tim.hansen@colostate.edu} \\
   \And
  Hossein Moradi Rekabdarkolaee\\
Business Analytics, Economics, and Information System \\
  Bowling Green State University,\\
   Bowling Green, Ohio\\
    \texttt{hmrekab@bgsu.edu} \\
}

\begin{document}
\maketitle

\begin{abstract}
Accurate forecasting and temporal downscaling of solar irradiance is critical for reliable renewable energy integration and grid management. Traditional methods often struggle to capture rapid irradiance variability. In this work, we develop a Transformer-based architecture for simultaneous forecasting and temporal downscaling. The model takes input data at arbitrary resolutions and generates outputs at any resolutions represented within the training data, leveraging both global and local contextual information. To improve training efficiency, a genetic algorithm-based optimization framework is implemented to identify suitable model hyperparameters, while reducing the computational cost of exhaustive grid search. The effectiveness of the proposed model is shown by analyzing real data in Dallas, Texas. The proposed model achieved root mean square error (RMSE) values of 28.785, 29.544, 30.157, 32.132~W/m\(^2\) for 15-, 30-, 60-minute, and irregularly spaced global input and producing forecast at 5-minutes resolution, respectively. Furthermore, this model produced reliable uncertainty estimates, exhibiting wider prediction intervals for high-resolution inputs during volatile solar conditions, whereas coarser-resolution inputs result in narrower intervals and significantly lower training time and computational cost. Our approach provides a scalable and efficient framework for multi-resolution solar irradiance forecasting and temporal downscaling, suitable for practical energy forecasting and grid management applications. 
\end{abstract}


\section{Introduction}
Rapid expansion of photovoltaic (PV) power has made solar energy a cornerstone of the global transition toward sustainable electricity generation \cite{wang2017outline, cameron2017solar}. As renewable energy and distributed energy resources (DERs) make up a larger share of electricity generation, keeping the grid stable and reliable becomes more challenging \cite{lew2013western, bright2017synthetic}. These challenges are increasingly important as DERs are integrated into bulk power system operations and wholesale electricity markets, including through regulatory efforts such as FERC Order No. 2222~\cite{ferc_order2222_2020}. 
Therefore, accurate solar irradiance forecasting is essential for PV power prediction, grid stability analysis, DER aggregation, energy market participation, and long-term resource assessment \cite{van2018probabilistic, dou2023day, sudharshan2022systematic}. 

A key challenge is that solar irradiance is influenced by multiple sources, making it volatile at high temporal resolutions. Consequently, temporal resolution plays a crucial role in solar irradiance modeling, as relying on hourly averaged data can bias PV system design and operational conclusions compared with high-frequency measurements \cite{mayer2021effects,cao2014impact}. 
This variability in solar irradiance requires forecasting models to capture both longer-term temporal structure and short-lived fluctuations \cite{camacho2025short}. 
To address these challenges, a wide range of solar irradiance forecasting methodologies have been developed. These methods can be classified into four main categories: numerical weather prediction models (NWP; \citealt{zhang2022solar}), image-based models \cite{mercier2024vision}, data-driven models \cite{obiora2023predicting}, and hybrid frameworks \cite{huang2021hybrid}. The NWPs remain widely used in application, however they are often computationally expensive and their accuracy can be sensitive to configuration choices and model-chain components \cite{dou2023day, mayer2021extensive}. Image-based forecasting methods can better capture rapid cloud evolution because sky images provide direct, high-resolution information on cloud changes, but they also face practical challenges in image acquisition, pre-processing, cloud tracking, and forecast-model design \cite{lin2023recent}. Data-driven forecasting methods can achieve high predictive accuracy by learning complex nonlinear and long-range dependencies directly from historical irradiance and meteorological data, but their performance is often constrained by substantial data requirements, high computational cost, and sensitivity to data quality and operational conditions \cite{jiao2025review}. Hybrid solar irradiance forecasting models aim to improve predictive accuracy by combining complementary techniques to handle the non-stationary and highly variable nature of solar radiation, but their greater architectural complexity can make model design, tuning, and deployment more challenging \cite{guermoui2020comprehensive}.

Recently, increasing attention has shifted toward attention-based Transformer architectures, which are designed to model long-range dependencies efficiently \cite{kumari2021deep, wen2022transformers, su2025systematic}. 
In renewable-energy contexts, Transformers models have shown strong performance for capturing complex spatiotemporal patterns \cite{hussan2026transformer}. Furthermore, Transformer-based models have shown to outperform recurrent and statistical baselines, particularly for longer horizons, and improve robustness under rapidly varying atmospheric conditions \cite{boussif2023improving, shan2025investigation, li2025transformer}. Despite 
these advances, two major issues remain for practical deployment: First, Transformer forecasters assume fixed input/output temporal resolutions, and second, performance is often highly sensitive to hyperparameter choices \cite{chowdhury2022comparative}. The fixed-resolution assumption becomes especially restrictive when measurements are available at heterogeneous temporal resolutions or when forecasts are required at a higher resolution than the input data \cite{deep2024deep}. In this paper, we propose a \emph{resolution-adaptive} Transformer framework for one-shot solar irradiance forecasting and downscaling. Our proposed model uses the global irradiance at arbitrary coarse resolutions (e.g., 15, 30, or 60 minutes, while also supporting irregular time series inputs) and 
generates downscaled high-resolution local forecasts (e.g., 5-minute intervals) for the next 24 hours. To reduce the computational burden, we integrate a genetic algorithm (GA) to efficiently search the model configuration space, using evolutionary optimization for large hyperparameter spaces \cite{waysi2025optimization}. The GAs naturally handle mixed-type search spaces, explore large, non-convex, and multi-modal landscapes; and reduce the risk of premature convergence by maintaining population diversity throughout the optimization process \cite{brzkek2025exploration}. Our approach provides a flexible forecasting-and-downscaling pipeline designed for heterogeneous-resolution data. To evaluate its performance against established forecasting approaches, the proposed framework is compared with four state-of-the-art Transformer-based models: Informer \cite{zhou2021informer}, Autoformer \cite{wu2021autoformer}, FEDformer \cite{zhou2022fedformer}, and the Non-stationary Transformer \cite{liu2022non}. These models were selected because they employ encoder--decoder architectures comparable to that of the proposed framework while using distinct mechanisms to capture temporal patterns in time-series data. Informer uses ProbSparse attention, Autoformer combines series decomposition with an Auto-Correlation mechanism, FEDformer models temporal patterns in the frequency domain, and the Non-stationary Transformer incorporates series stationarization and de-stationary attention.

The remainder of this paper is organized as follows. Section~\ref{sec:data} introduces the datasets and preprocessing procedures used in the study. The relevant Transformer architecture and GA concepts are reviewed in Section~\ref{sec:background}. Section \ref{sec:methodology} then develops the proposed forecasting model and optimization framework. Experimental results and performance comparisons are reported in Section~\ref{sec:results}. Finally, Section~\ref{sec:conclusions} summarizes the key findings and outlines future research directions. Additional results are provided in the Appendix.

\section{Data}
\label{sec:data}

This study utilizes two solar irradiance datasets: a global dataset obtained from the National Solar Radiation Database (NSRDB; \citealp{Sengupta2018NSRDB}) and a local ground-measured dataset \cite{AmbientWeather}. The global dataset provides Global Horizontal Irradiance (GHI) for Dallas, Texas, with \(4~\mathrm{km}^2\) gridded spatial resolution and a temporal resolution of 15-minutes, obtained from 1 January 2024 to 31 December 2024. The local dataset consists of 5-minute solar irradiance measurements collected at a weather station located at Camp David, Dallas, Texas, over the same time period. The goal is to downscale the coarser data using finer-resolution observations as reference and to simultaneously forecast future values at the finer scale. 

\begin{figure}
\centering
\includegraphics[width=\textwidth]{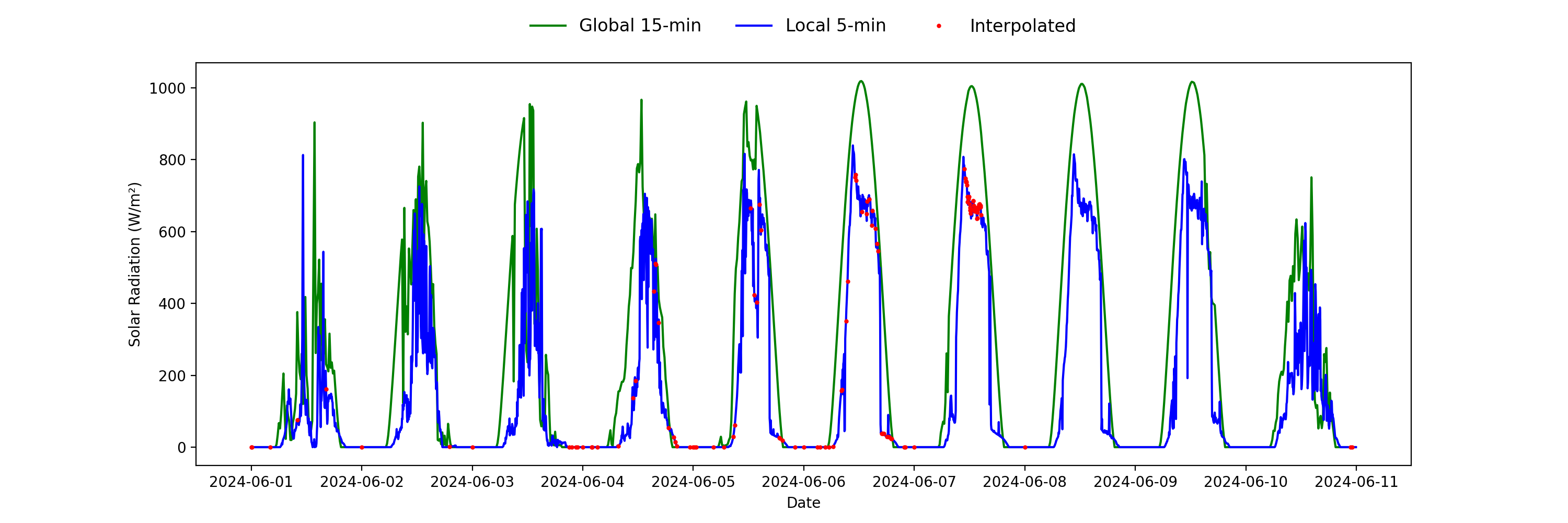}
\caption{\label{fig:dataset}Solar irradiance comparison between local 5-minute and global 15-minute datasets from June 1, 2024 to June 10, 2024. PCHIP interpolated local points for filling missing values are highlighted in red.}
\end{figure}

The global GHI dataset contains 35,136 observations, while the local solar irradiance dataset contains 103,744 observations. The global dataset had no missing observations. In contrast, the local dataset contained 12 duplicate timestamps and 1,910 missing observations, including gaps where the timestamps were not evenly spaced at 5-minute intervals. Duplicate timestamps were handled by retaining the first occurrence. The missing values were then imputed using Piecewise Cubic Hermite Interpolating Polynomial interpolation (PCHIP; \citealp{fritsch1980monotone}). The PCHIP is a shape-preserving interpolation technique that avoids overshooting and prevents the introduction of artificial oscillations between data points. This property is essential for solar irradiance data, where zero-irradiance periods (e.g., nighttime) must remain zero. A representative snapshot of the local 5-minute and global 15-minute solar irradiance datasets is shown in Figure~\ref{fig:dataset}, highlighting 
the presence of interpolated points in the local measurements for the period of June 1, 2024 to June 10, 2024. 

The nature of the data raises several challenges. First, the global and local irradiance measurements are recorded at different temporal resolutions, with global irradiance data available at 15-, 30-, 60-minute, and irregular intervals, while the local irradiance measurements are consistently recorded at a 5-minute resolution. This requires careful alignment of input and target pairs during pre-processing to maintain temporal continuity. Second, the local dataset contains missing observations, some occurring during daytime, where accurate interpolation is essential due to rapid irradiance fluctuations. Additionally, satellite-derived GHI tends to smooth high-frequency cloud-induced variability that local ground sensors captures, resulting in mismatches in both variability and amplitude between the local and global data. 

To address these challenges, the proposed framework leverages a Transformer architecture with cross-attention, which allows input and target sequences of different lengths. 
Missing local observations are filled using PCHIP interpolation, while Informer-style temporal embeddings provide time-related context that helps the model learn relationships between satellite-derived GHI and local ground observations despite differences in temporal resolution, variability, and magnitude. The data were used in their raw form without applying standardization or normalization.

\section{Brief Review} 
\label{sec:background}

\subsection{Transformers}

Transformers are attention-based architectures originally developed for sequence-to-sequence learning  \cite{vaswani2017attention}. Compared to traditional architectures such as Recurrent Neural Networks (RNNs; \citealp{schmidt2019recurrent}), Long Short-Term Memory networks (LSTMs; \citealp{hochreiter1997long}), and Convolutional Neural Networks (CNNs; \citealp{o2015introduction}), Transformers offer superior handling of long-range dependencies and irregular events through attention mechanisms and adaptive receptive fields. Furthermore, their parallel computation significantly improves training efficiency, and the architecture naturally supports the integration of exogenous features for complex, multi-horizon forecasting \cite{lim2021temporal, wen2022transformers, sanford2024transformers}. Further details on the attention mechanism,, computational complexity, and practical adjustments of Transformers for time series can be found in the Appendix.

\subsection{Hyperparameter Optimization via Genetic Algorithms} 

Hyperparameters such as learning rate, batch size, and network depth strongly influence model behavior, generalization, and computational efficiency; poor choices can cause underfitting, overfitting, or wasted resources, making optimal hyperparameter selection essential for validation performance \cite{shanthi2022genetic}. However, because hyperparameters cannot be optimized directly through gradient-based learning \cite{raiaan2024systematic}, and common methods such as grid search, random search, and Bayesian optimization have computational and theoretical limitations \cite{yu2020hyper}, increasingly complex models motivate more efficient alternative optimization strategies. In this paper, we use a Genetic Algorithm (GA), a population-based optimization method inspired by natural selection and commonly applied to hyperparameter tuning \cite{katoch2021review, mehdary2024hyperparameter}. The GA begins with a randomly sampled population of candidate hyperparameter sets, trains and evaluates each model using RMSE on a held-out test set, selects the top $n$ lowest-RMSE candidates at each generation, and iteratively applies crossover and mutation until RMSE convergence, with the final configuration chosen as the one achieving the lowest RMSE across all generations.







\section{Methodology}
\label{sec:methodology}
\subsection{Proposed Transformer}
\label{proposed_transformer}

\begin{figure}
\centering{
\includegraphics[width=0.5\textwidth]{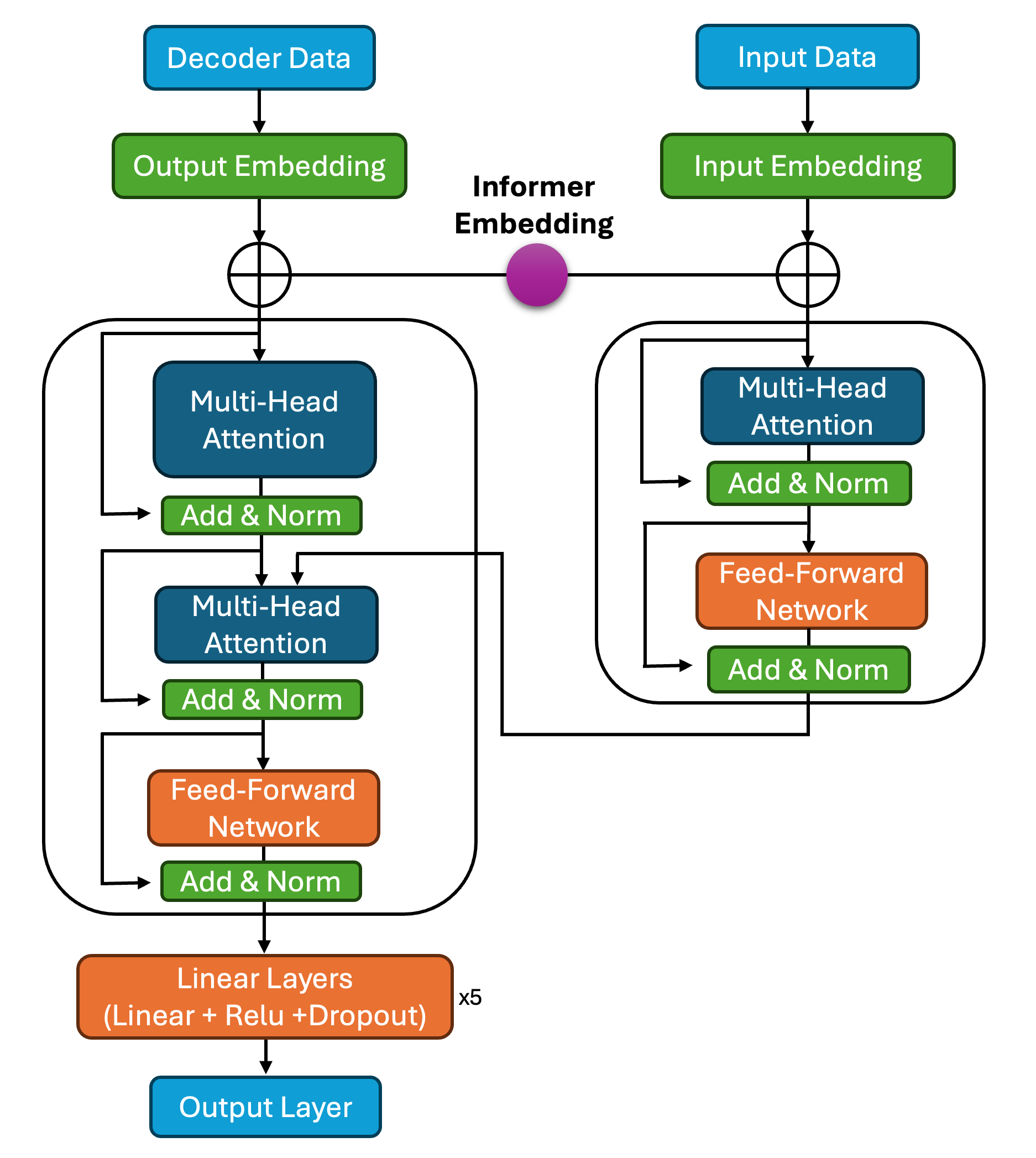}
}
\caption{\label{flowdiagram}Resolution-adaptive Transformer architecture for one-shot solar irradiance forecasting and downscaling.}
\end{figure}

The proposed Transformer model performs day-ahead forecasting and downscaling of the local irradiance measurments using the GHI (and potentially other features). An overview of the resolution-adaptive Transformer architecture is shown in Figure~\ref{flowdiagram}. For a batch of input-target pairs $(X,Y)$, let the global irradiance input tensor be denoted by 
\begin{equation}
X \in \mathbb{R}^{B \times T_{\mathrm{in}} \times f}, 
\label{eq:feature}
\end{equation}
where $B$ is the batch size, $T_{\mathrm{in}}$ is the input sequence length, and $f$ is the number of input features in the global dataset. The $i^{th}$ input sample in the batch is
\begin{equation*}
X_i = (\mathbf{x}_{i1}, \mathbf{x}_{i2}, \dots , \mathbf{x}_{iT_{\mathrm{in}}})
\in \mathbb{R}^{T_{\mathrm{in}} \times f} ; i=1,\dots,B ,
\label{eq:feature_sample}
\end{equation*}
where each $\mathbf{x}_{it} \in \mathbb{R}^{f}$ denotes the feature vector at time step $t$, containing all $f$ input features observed at that timestamp. In our case $f=1$ and $\mathbf{x}_{it} \in \mathbb{R}^{1}$, since we have only GHI as our input feature.

Similar to Informer-based architecture \cite{zhou2021informer}, the $X$ is projected into a higher-dimensional space using token, temporal, and positional embeddings before feeding it to the encoder. The process involves the following steps: 
\begin{enumerate}
\item \textbf{Token Embedding:} The input tensor $X \in \mathbb{R}^{B \times T_{\mathrm{in}} \times f}$ is first mapped into a $d$-dimensional representation using a one-dimensional convolution applied along the temporal axis. Specifically, the convolution uses $f$ input channels, $d$ output channels, kernel size $3$, circular padding, and stride $1$. Thus, each embedded token is obtained from a local temporal neighborhood around time step $t$ rather than from $\mathbf{x}_{it}$ alone:
\begin{equation*}
E^{\text{token}}(X) = \mathrm{Conv1D}(X) \in \mathbb{R}^{B \times T_{\mathrm{in}} \times d}.
\end{equation*}
Because padding is used, the temporal length is preserved, so the output has the same sequence length $T_{\mathrm{in}}$.
\item \textbf{Temporal Embedding:} To encode calendar information, a temporal feature tensor $X_{\mathrm{mark}} \in \mathbb{R}^{B \times T_{\mathrm{in}} \times 5}$ is constructed from the timestamps associated with the input sequence, where the five components at each time step are month (mon), day (d), weekday (w), hour (h), and minute (min). The $i$th sample in the batch is
\begin{equation*}
X_{\mathrm{mark},i} \in \mathbb{R}^{T_{\mathrm{in}} \times 5}, \qquad i=1,\dots,B,
\end{equation*}
with
\begin{equation*}
X_{\mathrm{mark},i} = \left( \mathbf{x}^{\mathrm{mark}}_{i1}, \mathbf{x}^{\mathrm{mark}}_{i2}, \dots, \mathbf{x}^{\mathrm{mark}}_{iT_{\mathrm{in}}} \right),
\end{equation*}
where $\mathbf{x}^{\mathrm{mark}}_{it} = [\mathrm{mon}_{it},\, \mathrm{d}_{it},\, \mathrm{w}_{it},\, \mathrm{h}_{it},\, \mathrm{min}_{it}]\in \mathbb{R}^{5}$, for $t=1,\dots,T_{\mathrm{in}}$. In our implementation, five separate fixed embedding layers are used for the temporal features month, day, weekday, hour, and minute. Each layer maps the corresponding integer calendar value to a $d$-dimensional vector whose values are learned jointly with the other model parameters during training. The resulting vectors are then summed to obtain the temporal embedding:
\begin{equation*}
E^{\mathrm{temp}}(\mathbf{x}^{\mathrm{mark}}_{it}) = E^{\mathrm{mon}}(\mathrm{mon}_{it}) + E^{\mathrm{d}}(\mathrm{d}_{it}) + E^{\mathrm{w}}(\mathrm{w}_{it}) + E^{\mathrm{h}}(\mathrm{h}_{it}) + E^{\mathrm{min}}(\mathrm{min}_{it}),
\end{equation*}
where $E^{\mathrm{temp}}(\mathbf{x}^{\mathrm{mark}}_{it}) \in \mathbb{R}^{d}$. Applying this across all samples and time steps gives the temporal embedding tensor $E^{\mathrm{temp}}(X_{\mathrm{mark}} ) \in \mathbb{R}^{B \times T_{\mathrm{in}} \times d}$. 


\item \textbf{Positional Embedding (PE):} A fixed sinusoidal PE is used to inject sequential information into the Transformer input, i.e. for $j = 0,\dots,(d/2)-1$, we have
\begin{equation*}
\mathrm{PE}_{t,2j} = \sin\left(\frac{t}{10000^{\frac{2j}{d}}}\right), \quad \mathrm{PE}_{t,2j+1} = \cos\left(\frac{t}{10000^{\frac{2j}{d}}}\right),
\end{equation*}
where $d$ is the embedding dimension. Instead of applying the same relative positional encoding to each sequence in the mini-batch, the implementation generated shifted positional encoding windows across the mini-batch. For an input batch of size $B$ and sequence length $T_{\mathrm{in}}$, the first batch row received positions $0,\dots,T_{\mathrm{in}}-1$, the second row received positions $1,\dots,T_{\mathrm{in}}$, and so on. Since mini-batches were shuffled during training, this shift corresponded to the row position within the mini-batch rather than the true temporal position of the original time-series window.
\end{enumerate}
The final combined embedding at time step $t$ for the $i$th sample is
\begin{equation*}
\mathrm{Embed}(\mathbf{x}_{it}) =E^{\text{token}}(\mathbf{x}_{it}) + E^{\text{temp}}(\mathbf{x}^{\mathrm{mark}}_{it}) + PE_t, \quad t = 1, \dots, T_{\mathrm{in}},
\end{equation*}
where $\mathrm{Embed}(\mathbf{x}_{it}) \in \mathbb{R}^{d}$. When the full input tensor $X \in \mathbb{R}^{B \times T_{\mathrm{in}} \times f}$ is passed through the final embedding layer (Embed(.)), the input feature vector is mapped to a $d$-dimensional representation, $E = \mathrm{Embed}(X)\in \mathbb{R}^{B \times T_{\mathrm{in}} \times d}$.

The \textbf{encoder} consists of $L$ stacked layers, each with multi-head self-attention (MHA), a ReLU-based feed-forward network (FFN), and residual connections. For layer $l = 1, \dots, L$, the computations are given by
\begin{eqnarray}
\label{eq1}
\tilde{H}^{(l)}&=& \mathrm{LayerNorm}\!\left(H^{(l-1)} + \mathrm{MHA}\!\left(H^{(l-1)}\right) \right),\cr
H^{(l)} &=& \mathrm{LayerNorm}\!\left(\tilde{H}^{(l)} +   \mathrm{FFN}\!\left(\tilde{H}^{(l)}\right) \right),
\end{eqnarray}
where LayerNorm(.) denotes the layer normalization. In Equation (\ref{eq1}), $H^{(0)} = E$ denotes the embedded input sequence. The encoder output after final layer is $H^{(L)} \in \mathbb{R}^{B \times T_{\mathrm{in}} \times d}$.

Let $Y \in \mathbb{R}^{B \times T_{\mathrm{out}} \times 1}$ denote the target tensor where $B$ is the batch size and $T_{\mathrm{out}}$ is the target sequence length. To construct the decoder input for one-shot forecasting, the first $T_{\mathrm{label}}$ time steps of $Y$ are retained as observed values (denoted by $Y_{\mathrm{label}}$), where $T_{\mathrm{label}} \leq T_{\mathrm{out}}$. The remaining $T_{\mathrm{forecast}} = T_{\mathrm{out}} - T_{\mathrm{label}}$ positions are filled by repeating the last observed value in $Y_{\mathrm{label}}$ (denoted by $Y_{\mathrm{pad}}$). Accordingly, the observed part is $Y_{\mathrm{label}} = Y_{[:,\,0\,:\,T_{\mathrm{label}},\::]} \in \mathbb{R}^{B \times T_{\mathrm{label}} \times 1}$, and the padded part is $Y_{\mathrm{pad}} = \underbrace{\big[Y_{[:,\,T_{\mathrm{label}},\: :]}, \dots, Y_{[:,\,T_{\mathrm{label}},\:: ]}\big]}_{T_{\mathrm{forecast}} \text{ times}} \in \mathbb{R}^{B \times T_{\mathrm{forecast}} \times 1}$.

The \textbf{decoder} input is then formed by concatenating these two parts:
\begin{equation*}
Y_{\mathrm{decoder}} = \mathrm{concat}(Y_{\mathrm{label}},\, Y_{\mathrm{pad}})
\in \mathbb{R}^{B \times T_{\mathrm{out}} \times 1}.
\end{equation*}
The decoder input $Y_{\mathrm{decoder}}$ is subsequently mapped into the same $d$-dimensional embedding space as the encoder input using the embedding function $\mathrm{Embed}(\cdot)$,
\begin{equation*}
D = \mathrm{Embed}\!\left(Y_{\text{decoder}}\right) \in \mathbb{R}^{B \times T_{\text{out}} \times d}.
\end{equation*}
The decoder consists of $L$ stacked Transformer decoder layers. Each layer includes a multi-head self-attention sublayer ($\mathrm{MHA}_{\text{self}}$), a cross-attention sublayer ($\mathrm{MHA}_{\text{cross}}$), and a ReLU-based feed-forward network (FFN). Residual connections and layer normalization are applied after each sublayer. For layer $l = 1, \dots, L$, the computations are given by
\begin{eqnarray*}
\tilde{D}^{(l)} &=& \mathrm{LayerNorm}\!\left(D^{(l-1)} + \mathrm{MHA}_{\text{self}}\!\left(D^{(l-1)}\right) \right),\cr
\hat{D}^{(l)} &=& \mathrm{LayerNorm}\!\left(\tilde{D}^{(l)} +\mathrm{MHA}_{\text{cross}}\!\left(\tilde{D}^{(l)},\, H^{(L)}\right) \right), \cr
D^{(l)} &=& \mathrm{LayerNorm}\!\left( \hat{D}^{(l)} + \mathrm{FFN}\!\left(\hat{D}^{(l)}\right) \right),
\end{eqnarray*}
where $D^{(0)}=D$. The cross-attention mechanism enables the decoder to attend to the encoder representations, which is central to the downscaling task. In cross-attention, the decoder hidden states provide the query representations, while the encoder outputs supply the corresponding key and value representations. By attending to encoder-derived keys and values computed from coarser-resolution GHI, and processing decoder queries derived from finer-resolution local irradiance, the model learns cross-scale dependencies without explicit resampling or manual temporal alignment. After the final decoder layer, the output has dimensions $D^{(L)} \in \mathbb{R}^{B \times T_{\text{out}} \times d}$. No causal masking is applied in the decoder self-attention. Unlike autoregressive approaches, predictions are generated simultaneously rather than stepwise (one-shot forecasting), making masking unnecessary and allowing the model to capture full-sequence dependencies more efficiently.

To capture fine-grained variations in the local irradiance, the decoder output is further processed through a stack of feed-forward layers arranged in a funnel-shaped architecture. The decoder output  $D^{(L)}$, is first reshaped by merging the batch and temporal dimensions: $Z^{(0)} = \mathrm{Reshape}\!\left(D^{(L)}\right) \in \mathbb{R}^{(B \cdot T_{\text{decoder}}) \times d}$. Then, it is passed through a sequence of six fully connected layers with dimensions: 1024, 512, 256, 128, 64 and 1. Each layer has ReLU activation enabling the model to learn non-linear patterns in the forecasted sequence. The final layer corresponds to the predicted local solar irradiance. The output is then reshaped back to recover the original batch and temporal structure, yielding a tensor of shape $B \times T_{\text{out}} \times 1$, where $B$ is the batch size and $T_\mathrm{out}$ is the target sequence length. The last time steps of $T_\mathrm{out}$, denoted by $T_{\mathrm{}{forecast}}$, are retained as the final forecast. 

\subsection{Model Training and Computational Environment}
The model is trained by minimizing the mean squared error (MSE) on the training set and using the AdamW optimizer for optimization. To improve training stability and convergence, we employ the \texttt{ReduceLROnPlateau} learning rate scheduler, which adaptively reduces the learning rate when the training loss ceases to improve. During training, dropout with a rate of 0.1 is applied uniformly across the encoder, decoder, and embedding layers to mitigate overfitting. Core architectural components of the Transformer are treated as tunable hyperparameters. In addition, parameters such as input sequence length, batch size, learning rate, and decoder label length are also optimized.

The GA algorithm is employed to obtain the best hyperparameter combination. A selection rate of 0.5 is used, meaning that parents are chosen from the top 50\% of individuals in each generation. Offspring are generated using a crossover probability of 0.8, while a mutation rate of 0.2 introduces random mixing to maintain population diversity. In addition, an elitism rate of 0.2 ensures that a fraction of the best-performing individuals is preserved across generations, helping to retain high-quality solutions throughout the evolutionary process. During the GA search, all candidate models are trained for a 100 epochs, which substantially reduces computational cost while allowing reliable comparison across the large hyperparameter space. The best-performing hyperparameter configuration is selected based on the minimum RMSE on the test set. The corresponding best models are re-trained for 300 epochs using the same hyperparameters and identical initialization conditions. The final 300 epoch rerun is to assess whether extended training yields additional performance improvements beyond those observed during the proxy training stage. 

All models are implemented in PyTorch and trained on a GPU-enabled cluster node equipped with an Intel Xeon Gold 6342 CPU @ 2.80\,GHz, up to 200\,GB of system memory, and an NVIDIA A100 GPU with 80\,GB of memory. CUDA version 11.8 is used for acceleration, with up to 15 CPU cores allocated for data loading and preprocessing.

\subsection{Uncertainty Estimates}
To evaluate the robustness and consistency of the proposed  model, we conducted a series of retraining experiments for each of the best-performing models corresponding to the four input-resolution scenarios. Specifically, each of the best model is retrained hundred times using the same optimal hyperparameter configuration identified by the genetic algorithm, while allowing all sources of randomness to vary freely across runs. This analysis isolates the variability in predictive performance attributable solely to random training effects. The uncertainty estimates are then constructed using two complementary approaches:
\begin{enumerate}
\item \textbf{Performance-based uncertainty bounds:} After retraining the selected best model 100 times, the retrained models are ranked based on their RMSE values. The predictions from the 3rd-best and 97th-best retrained models are then selected and plotted along with the prediction from the best-performing model. 
\item \textbf{Pointwise variance-based uncertainty estimates:} Since each 24-hour forecast consists of 288 prediction points at 5-minute intervals, each forecast time step has 100 predicted values obtained from the 100 independent retraining runs. For each time step $t$, where $t = 1,\ldots,288$, a pointwise sample variance is computed across the 100 predictions. However, instead of centering the variance around the mean prediction at time step $t$, the variance is computed relative to the prediction from the best-performing model at the same time step. Specifically, if $\hat{y}_{r,t}$ denotes the prediction from retraining run $r$ at time step $t$, and $\hat{y}_{\text{best},t}$ denotes the prediction from the best-performing model (from the GA trained models) at the same time step, the pointwise variance is calculated as
\[
s_t^2 = \frac{1}{n-1} \sum_{r=1}^{n} \left(\hat{y}_{r,t} - \hat{y}_{\text{best},t}\right)^2, \qquad n = 100.
\]
This allows us to capture timestep-specific variability in the model outputs.
\end{enumerate}

\subsection{Benchmark Models and Comparative Setup}
\label{sec:benchmark_models}

To evaluate the proposed framework against established forecasting methods, four Transformer-based encoder--decoder models were implemented as benchmarks: Informer, Autoformer, FEDformer, and the Non-stationary Transformer. Custom PyTorch implementations were constructed following the main architectural components of the original models and were adapted to the same global-to-local forecasting and downscaling setting used by the proposed framework.
 
To ensure a consistent comparison, all benchmark models were trained and optimized using the same loss function, optimizer, learning-rate scheduler, GA settings, and number of training epochs as the proposed Transformer framework.
In addition to the common hyperparameters, model-specific hyperparameters were included in the search space where required to account for the distinct architectural components of each benchmark model.

The principal model-specific components were retained in each implementation. Informer used ProbSparse self-attention, encoder distillation, and full cross-attention in the decoder. Autoformer used progressive seasonal--trend decomposition and the Auto-Correlation mechanism. Because the encoder and decoder operate on different irradiance series, the Autoformer decomposition was initialized using the known local decoder history rather than the global encoder series. FEDformer followed a similar decomposition-based structure but replaced the Auto-Correlation mechanism with Fourier-domain self- and cross-attention. Its decomposition was also initialized from the known local history, and the GA factor parameter was mapped to the number of Fourier modes. The Non-stationary Transformer applied series stationarization and de-stationary attention, with the scaling and shifting factors learned from the raw global encoder sequence. Its predictions were restored using statistics from the local decoder history because the forecasting target was local irradiance. In Autoformer and FEDformer, boundary values were repeated only within the moving-average decomposition to preserve the original sequence length. For the Non-stationary Transformer, the learned attention-shift vector was padded with zeros or truncated when necessary to match the encoder or decoder key length. These operations were internal architectural adjustments and did not modify the input datasets.

\section{Results and Discussion}
\label{sec:results}

The complete dataset spans the full year of 2024. Two evaluation settings were considered: (1) a controlled comparison of four model variants using the May--June subset and (2) a rolling monthly evaluation of the 60-minute input model across the full year.

For the controlled comparison, four variants of the proposed model were trained using global GHI inputs with different temporal sampling characteristics: 15-minute GHI (\textbf{Model 1}), 30-minute GHI (\textbf{Model 2}), 60-minute GHI (\textbf{Model 3}), and irregularly sampled GHI (\textbf{Model 4}). The 30-minute and 60-minute datasets were derived from the original 15-minute global GHI series by retaining every second and fourth observation, respectively. To construct the irregular-input case, observations were randomly removed without replacement from the 15-minute May--June series. The original series contained 5,856 observations, of which 3,800 were removed, leaving 2,056 observations with irregular temporal spacing.

All four models generate local solar irradiance forecasts at a fixed 5-minute resolution using a one-shot, full-day forecasting strategy. Although the output resolution is fixed at 5 minutes in the present study, the proposed architecture can be adapted to generate forecasts at other temporal resolutions.

The June 30, 2024 forecasting scenario was used for the direct comparison among the four model variants. Because the input datasets have different sampling intervals, the exact 24-hour output windows differ slightly among the models:
\begin{itemize}
\item \textbf{Model 1}: June 29, 2024 at 23:50 to June 30, 2024 at 23:45,
\item \textbf{Model 2}: June 29, 2024 at 23:35 to June 30, 2024 at 23:30,
\item \textbf{Model 3}: June 29, 2024 at 23:05 to June 30, 2024 at 23:00, and
\item \textbf{Model 4}: June 29, 2024 at 23:50 to June 30, 2024 at 23:45.
\end{itemize}

To assess performance across different seasonal conditions, an additional full-year evaluation was conducted using only the 60-minute input configuration (\textbf{Model 3}). A rolling monthly prediction scheme was applied in which the final day of each month served as the testing period and the preceding training window extended over approximately two months, ending on the day before the test date. For example, the prediction for April 30, 2024 was generated using observations from March 1 through April 29, 2024 for training. For January, because earlier observations were unavailable, the model was trained using data from January 1 through January 30 and evaluated on January 31. The monthly forecasting results for Model 3 are provided in the Appendix.

\subsection{Training and Evaluation Setting}
\label{training_eval_setting}

For the controlled comparison, the dataset was split chronologically into a training period from May 1 to June 29, 2024, and a testing period consisting of June 30, 2024.  The solar irradiance time series exhibits short-range dependence. Hence we do not use a separate validation split because removing a continuous segment would break the temporal continuity most relevant for forecasting the test day. To construct training samples, we generate paired sequences of GHI inputs and corresponding local irradiance targets. The global input and local target sequences are temporally aligned to ensure continuity across the different input resolutions. Each sample is defined by three parameters: the input sequence length ($T_{\text{in}}$), the label length ($T_{\text{label}}$), and the forecasting horizon ($T_{\text{forecast}}$). 

For each scenario, The GA searches over the following ranges: input sequence length from 48 to 384 time steps, embedding dimension from 16 to 512, number of attention heads from 1 to 16, and feed-forward network dimension from 64 to 2048. The number of encoder and decoder layers is varied from 1 to 4, batch size from 16 to 256, learning rate from $10^{-6}$ to $10^{-2}$, and decoder label length from 24 to 144 time steps. The GA runs for 10 generations with a population size of 100, resulting in approximately 1000 models evaluated per resolution. 
These settings were chosen based on the available computational resources and the observed convergence of the search, as the later generations did not produce solutions that outperformed the best configuration identified earlier. Table~\ref{Best Models Obtained} summarizes the GA results.

For consistency, the benchmark models were evaluated under the same controlled May--June setting used for Models 1--4. All models were trained and tested using the same preprocessed datasets, chronological data split, calendar features, and timestamp-alignment procedure. This common experimental setup ensured that the observed performance differences were primarily attributable to the forecasting architectures rather than variations in data preparation.

\subsection{Forecasting Results}
Table~\ref{Hyperparameters} shows the hyperparameter configurations corresponding to these best-performing models as well as the total training time to complete the GA process for each resolution scenario. The total computational training time was greatest for Model 1 and lowest for Model 3. In specific, Model 1 required 10 days, 9 hours, 24 minutes, and 32 seconds to complete the full genetic algorithm search. In comparison, Model 2 and Model 3 required 2 days, 21 hours, 22 minutes, and 35 seconds, and 1 day, 16 hours, 26 minutes, and 50 seconds, respectively. Best architecture for Model 1 also has the largest sequence length, batch size, and learning rate. Models 2 and 3 have larger embedding size and larger or equal dimension for feed-forward neural network. These patterns are attributable to differences in the size of the input data across model configurations. 

\begin{table}[ht]
\caption{The hyperparameters of the best model and total time to complete the full GA search for each scenario. The training time is in days (d)/hours (h)/minutes (m)/seconds. The input resolution is 15 minutes, 30 minutes, 60 minutes, and irregular time series for Models~1,~2,~3, and~4, respectively, with day ahead forecast at 5 minutes resolution.}
\label{Hyperparameters}
\centering
\scalebox{0.9}{
\begin{tabular}{lllll}
\hline
\textbf{Hyperparameter} & \textbf{Model 1} & \textbf{Model 2} & \textbf{Model 3} & \textbf{Model 4} \\
\hline
Sequence Length      & 144   & 48    & 96 & 144 \\
Embedding Size       & 32    & 256   & 256 & 128\\
Attention Heads      & 16    & 16    & 16  & 4\\
Feed-forward Dim     & 512   & 1024  & 512 & 256\\
Number of Layers     & 4     & 3     & 4   & 2\\
Batch Size           & 128   & 32    & 96 & 16 \\
Learning Rate        & 0.001 & 0.0001 & 0.0001 &0.0001 \\
Label Length         & 24    & 24    & 24 & 144 \\
\hline
Total Training Time & 10d/9h/24m/32s & 2d/21h/22m/35s & 1d/16h/26m/50s & 2d/10h/38m/37s \\
\hline
\end{tabular}}
\end{table}

The extended training with 300 epochs did not yield performance gains. Consequently, the 100-epoch training configuration used during the GA search was retained as the final training setup for all four scenarios. Table~\ref{Best Models Obtained} summarizes the performance of the best models in each scenario. Among the four scenarios, the model trained with 15-minute GHI inputs achieves the lowest RMSE. This outcome aligns with expectations, as the finer-resolution global input provides richer temporal information, enabling the model to better capture short-scale variations in local solar irradiance \cite{zhang2018stochastic}. Similar to total training time, the computational training time for Model 1 is the longest and Model 3 required shorter training time for best architecture. As seen in Table~\ref{Hyperparameters}, although the difference in runtime for a single model run was approximately 10 minutes, the cumulative effect was substantial because the full training procedure required approximately 1000 model runs for each scenario. Model 1 produces smallest residual variance and converges in the earlier GA generation. Therefore, the choice of model depends on the desired trade-off between accuracy, computational time, and memory requirements. Model 1 is preferable when predictive accuracy is the primary goal, whereas Model 3 is advantageous when computational efficiency and reduced memory use are prioritized. Model 2 represents a compromise, offering reduced computational cost while preserving accuracy closer to that of Model 1.

For the irregular time series scenario, the training time for the best model was 4.07 minutes. Furthermore, the full genetic algorithm search required 2 days, 10 hours, 38 minutes, and 37 seconds to complete. These times are slightly greater than those observed in the 60-minute input scenario but remain lower than the other configurations. This may be attributable to the reduced number of available training points resulting from missing time steps. Notably, the presence of missing values was associated with the highest RMSE and residual variance across all scenarios. This degradation in predictive performance is consistent with prior deep learning time-series studies, which have shown that missing observations disrupt temporal dependencies, reduce the effective information available for model training, and can negatively affect downstream forecasting accuracy unless appropriately handled through imputation or missingness-aware modeling approaches \cite{fang2020timeseriesdataimputation, CiniMA22, wang2024deep}. These findings highlight the importance of complete and regularly sampled information for accurate forecasting.

\begin{table}
\caption{The RMSE, training and prediction time, variance of residual and generation of the Genetic Algorithm (GA Gen.) that produced the best results for each scenario. Input resolution is 15 minutes, 30 minutes, 60 minutes, and irregular time series for Models~1,~2,~3, and~4, respectively, with day ahead forecast at 5 minutes resolution.
The random seeds are 1316889585, 3794230437, 1186702747 and 2359100063 for Models~1,~2,~3, and~4, respectively.}
\label{Best Models Obtained}
\centering
\begin{tabular}{|c|c|c|c|c|c|}
\hline
\textbf{Model} & \textbf{RMSE} & \textbf{Train. Time} & \textbf{Pred. Time} & \textbf{Residual Variance} & {\textbf{GA Gen.}} \\ 
\hline
Model 1 & \textbf{28.785}  & 16.60 min & 0.0119 sec & \textbf{829.580} & 4 \\
Model 2 & 29.544 & 6.52 min & 0.0074 sec & 846.098 & 9\\
Model 3 & 30.157 & \textbf{3.14 min} & 0.009 sec &  898.916 & 7\\
Model 4 & 32.132 & 4.07 min & \textbf{0.0054} sec &  1025.702 & 6\\
\hline
\end{tabular}
\end{table}

\begin{figure}
\centering\includegraphics[clip,width=0.85\textwidth]{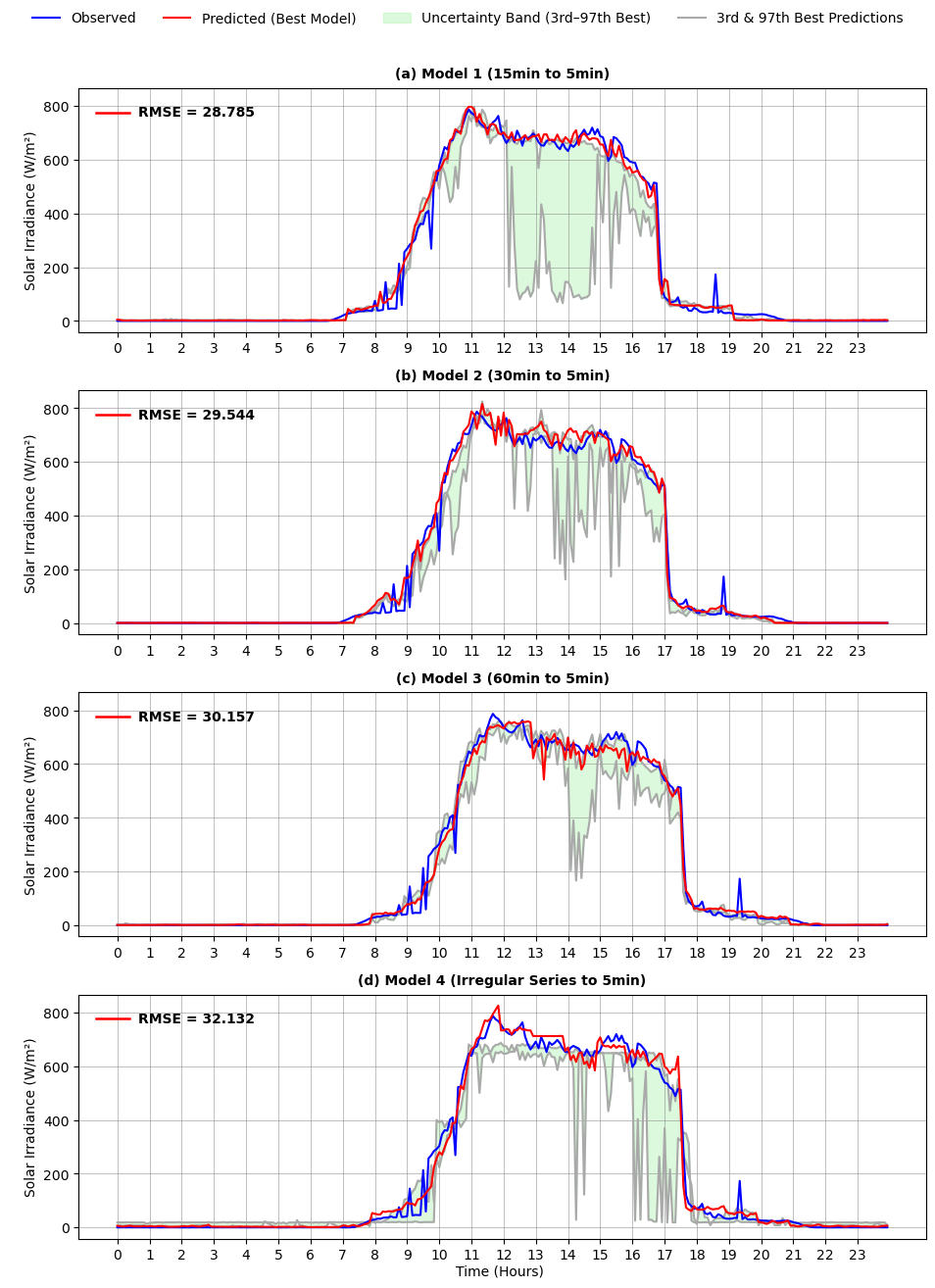}
\caption{Observed and predicted 5-minute local solar irradiance at Camp Davis, Dallas, from June~29 to June~30, 2024, using GHI inputs at 15-, 30-, 60-minute, and irregular temporal resolutions. Blue lines indicate observed irradiance, red lines indicate best-model predictions, and the shaded region represents the uncertainty band between the 3rd-best and 97th-best retrained models.}
\label{Fig:Best_Predictions}
\end{figure}

Figure~\ref{Fig:Best_Predictions} illustrates the performance of the each 
model and highlights the trade-off between input temporal resolution and forecasting accuracy. The observed local irradiance is shown in blue, while the corresponding best-model predictions are shown in red. To assess model consistency under random initialization the grey curves represent the 3rd-best and 97th-best retrained models, ranked according to RMSE, among the 100 independent retraining runs, with the light green shaded region indicating the variability between these bounds. All models successfully capture the overall pattern of solar irradiance. They closely track the low-variability periods such as sunrise and sunset. The largest discrepancies occur during peak solar hours, 10:00 to 18:00, when irradiance exhibits rapid fluctuations. Although Model~1 achieves the lowest RMSE, it still exhibits noticeable errors in tracking sharp dips and spikes. Nevertheless, its predictions follow the observed trends more closely than those of Models~2,~3 and~4. Across all scenarios, the models exhibit a tendency to smooth high-frequency variations, reflecting a common limitation in capturing instantaneous irradiance variability during temporal downscaling \cite{zhang2018stochastic}.

\subsection{Uncertainty Evaluation}

The prediction intervals, visualized in Figure~\ref{Fig:Best_Predictions}, remain narrow during early morning and late evening hours, indicating high confidence when irradiance is low or smoothly varying, but widen substantially during peak solar hours, reflecting increased uncertainty under highly variable conditions. The longest period for wide interval belongs to Model 1. 
The irregular-input scenario, Model 4, does not accurately capture the morning irradiance pattern and produces a wider range of predictions in the late afternoon. Nevertheless, its best-performing configuration mimics the overall temporal pattern of the local irradiance observations.

\begin{figure}
\centering\includegraphics[clip,width=0.85
\textwidth]{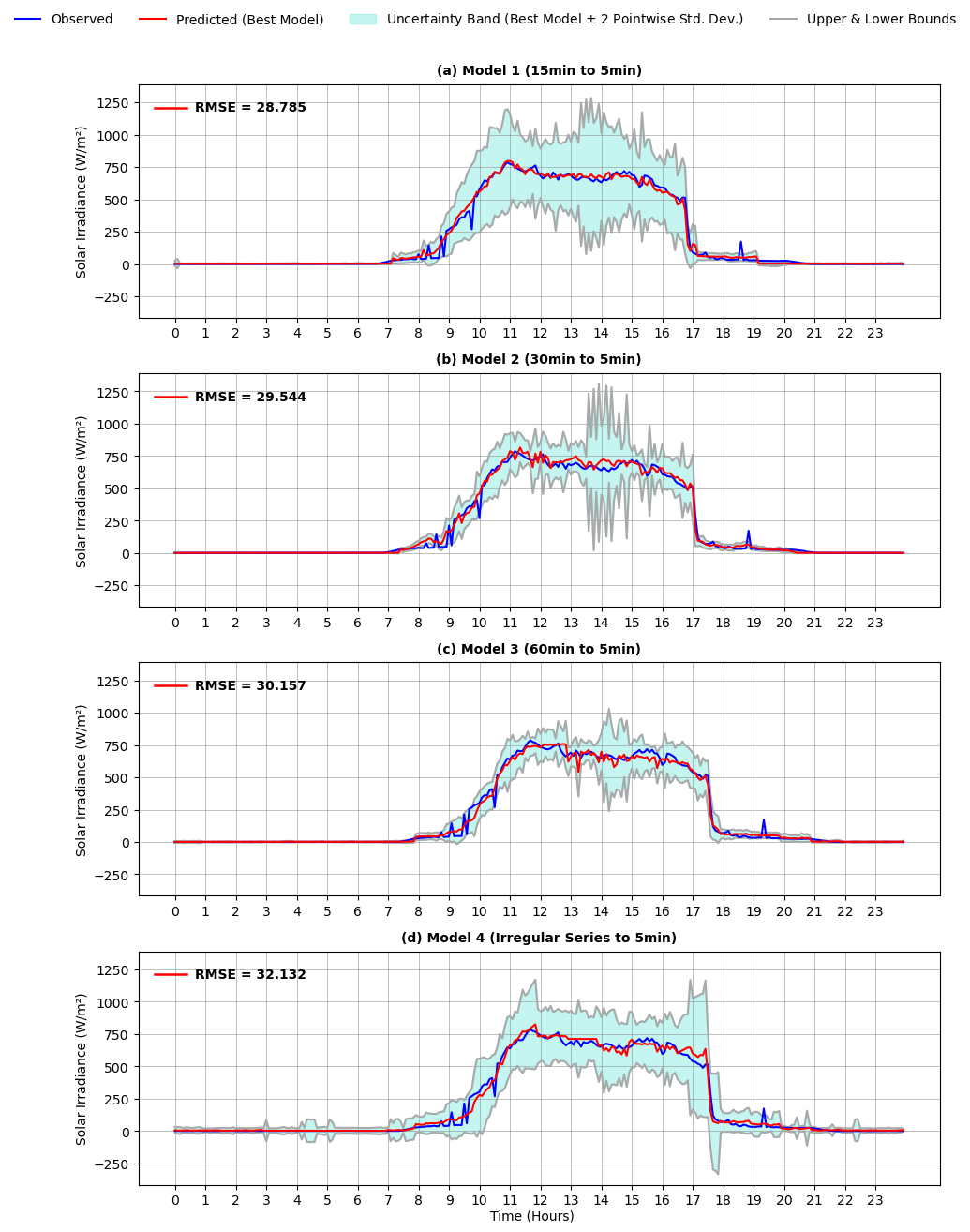}
\caption{Uncertainty quantification for predicted local solar irradiance at Camp Davis, Dallas, over a 24-hour period (June 29–30, 2024). The red line shows the best model prediction and the blue line represents the observed irradiance. The shaded region represents the uncertainty interval (best model $\pm 2$ pointwise standard deviations), quantifying prediction variability due to randomness in the training process at each forecast timestep.}
\label{Fig:Pointwise_Best_Predictions}
\end{figure}

To further assess the impact of stochastic training on forecast stability, we also compute a pointwise measure of variability for each 24-hour forecast horizon for the best models obtained in Table~\ref{Best Models Obtained}. 
For each forecast time step, the pointwise mean squared deviation from the best-model prediction was computed across the 100 independently retrained models. Figure~\ref{Fig:Pointwise_Best_Predictions} show the best-model forecast with an uncertainty interval defined by $\pm 2$ pointwise standard deviations. For Models 1 and 4, noticeable fluctuations appear in the uncertainty bounds during early morning and nighttime hours, when solar irradiance should be approximately zero. The lower bound also occasionally falls below zero, which is not physically meaningful because irradiance cannot be negative. These fluctuations reflect stochastic variability among repeated retraining runs under near-zero irradiance conditions, rather than real solar irradiance behavior. It is also observed that the models with coarser input resolution, Models 2 and 3, have narrower prediction interval compared to the finer input resolution model.

\subsection{Comparison with State-of-the-Art Forecasting Models}
\label{sec:sota_comparison}

For a consistent comparison and to limit computational cost, all benchmark models were evaluated using the 60-minute-to-5-minute forecasting setting. Each model received 60-minute global GHI as the encoder input and generated a one-shot 24-hour local irradiance forecast at 5-minute resolution. Table~\ref{tab:sota_comparison} compares the proposed RATSIFT model with the Non-stationary Transformer, FEDformer, Autoformer, and Informer. Lower values indicate better performance for RMSE, MAE, prediction time, training time, and residual variance.

\begin{table*}[!ht]
\centering
\small
\caption{Performance comparison of RATSIFT and the benchmark Transformer models. 
The best value in each performance metric is shown in bold.}
\label{tab:sota_comparison}
\begin{tabular}{lrrrrrr}
\toprule
\textbf{Model} &
\textbf{RMSE} &
\textbf{MAE} &
\textbf{Pred. Time} &
\textbf{Train. Time} &
\textbf{Residual Var.} &
\textbf{GA Gen.} \\
\midrule
RATSIFT & \textbf{30.157} & \textbf{17.119} & \textbf{0.0090 sec} & \textbf{3.14 min} & \textbf{898.916} & 7 \\
Non-stationary & 43.422 & 31.281 & 0.0243 sec & 6.40 min & 1793.040 & 10 \\
FEDformer & 46.979 & 32.771 & 0.1352 sec & 43.98 min & 2212.658 & 7 \\
Autoformer & 60.732 & 46.913 & 0.0268 sec & 13.14 min & 3657.485 & 10 \\
Informer & 189.902 & 113.397 & 0.0347 sec & 14.27 min & 24307.488 & 9 \\
\bottomrule
\end{tabular}
\end{table*}

RATSIFT achieved the lowest RMSE, MAE, and residual variance, indicating both greater forecasting accuracy and lower dispersion of the prediction errors. It was also the most computationally efficient model, requiring only 3.14 minutes for training and 0.009 seconds for prediction. The Non-stationary Transformer produced the second-best results, followed by FEDformer and Autoformer, whereas Informer exhibited substantially larger errors.

\begin{figure}
    \centering
    \includegraphics[width=1\linewidth]{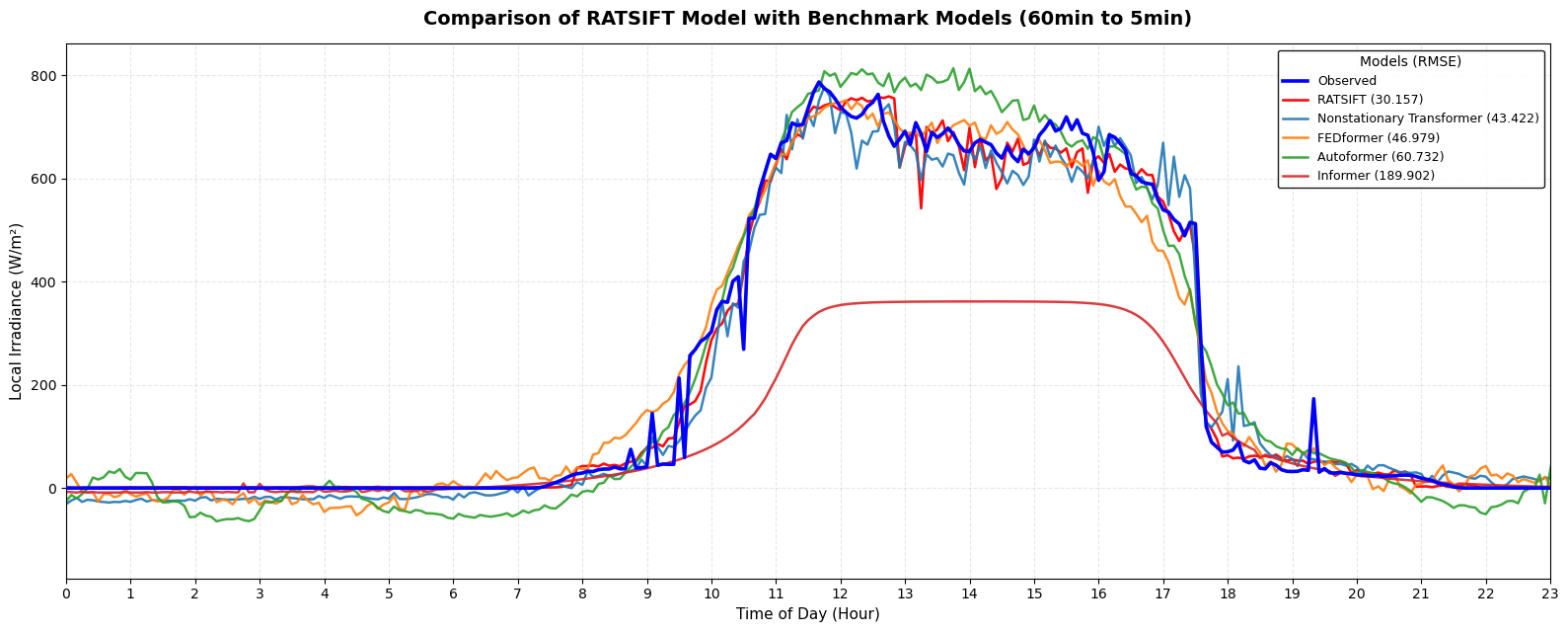}
    \caption{Observed and predicted 5-minute local solar irradiance profiles for June 30, 2024, obtained from 60-minute global GHI inputs. The proposed RATSIFT model is compared with the Non-stationary Transformer, FEDformer, Autoformer, and Informer, with the corresponding RMSE values reported in the legend.}
    \label{fig:benchmarks}
\end{figure}

Figure~\ref{fig:benchmarks} further illustrates the differences among the models. RATSIFT most closely reproduced the observed daily irradiance profile, including the morning increase, midday variation, and rapid evening decline. The Non-stationary Transformer and FEDformer captured the general daily pattern but showed larger deviations during the afternoon and near sunset. Autoformer tended to overestimate midday irradiance and produced unstable predictions during nighttime hours. Informer generated an overly smooth forecast and substantially underestimated the daytime irradiance magnitude, resulting in the largest forecasting errors.

The improved performance of RATSIFT may be attributed to its direct cross-resolution encoder--decoder formulation, which learns the relationship between coarse global GHI and fine-resolution local irradiance without imposing the explicit seasonal--trend decomposition used by Autoformer and FEDformer or the stationarization and de-stationarization framework used by the Non-stationary Transformer \cite{wu2021autoformer,zhou2022fedformer,liu2022non}. This flexibility appears to help the model reproduce the morning increase, short-term daytime variability, and evening decline more accurately.

\section{Conclusion}
\label{sec:conclusions}

This study introduced a resolution-adaptive Transformer framework for one-shot forecasting and downscaling of solar irradiance, with model configurations optimized using a genetic algorithm. The proposed framework uses global horizontal irradiance at multiple temporal resolutions as input and generates high-resolution 5-minute local irradiance forecasts. The results demonstrate that coarser-resolution inputs can achieve competitive predictive accuracy while substantially reducing computational cost. The GA-based hyperparameter optimization revealed a clear trade-off between temporal resolution and computational efficiency. Models trained on coarser-resolution inputs required significantly less optimization and training time, enabling the use of larger training datasets. 
In contrast, finer-resolution inputs retained greater temporal detail and produced marginally more accurate forecasts, consistent with previous research showing that substantial irradiance variability occurs at finer temporal scales \cite{zhang2018stochastic}, but at the expense of increased computational overhead. 

Repeated retraining experiments further showed that the framework produced reasonably accurate predictions and reliable uncertainty estimates. Prediction intervals remained narrow during stable irradiance periods and widened during highly variable peak solar hours, reflecting increased forecast uncertainty under rapidly changing conditions. The framework also performed effectively when the input consisted of an irregularly sampled time series. The proposed framework also outperformed the four benchmark Transformer architectures.

Overall, the proposed resolution-adaptive Transformer offers a scalable, flexible, and computationally efficient solution for multi-resolution solar irradiance forecasting and downscaling. Its ability to balance accuracy and efficiency makes the proposed approach well suited for practical applications such as photovoltaic power forecasting and grid operations, where accurate forecasts are important for managing spatially distributed solar generation \cite{visser2022operational}. Future work will extend the framework to spatial and spatiotemporal settings using multi-site and/or satellite data.  Previous studies have demonstrated the value of incorporating spatial context, geographically distributed stations, and satellite information into solar irradiance forecasting models \cite{boussif2023improving,li2025transformer}. Additional architectural enhancements may also improve robustness under highly variable weather conditions while preserving computational efficiency. Further improvements to the hyperparameter optimization process will be explored, including Bayesian optimization \cite{snoek2012practical}, Latin hypercube sampling for GA population initialization \cite{preechakul2009modified}, and adaptive crossover strategies that adjust genetic-operator probabilities according to model fitness \cite{srinivas1994adaptive}. These extensions are currently under investigation.

\section{Disclosure statement}\label{disclosure-statement}

The authors have no conflicts of interest.

\section{Data Availability Statement}\label{data-availability-statement}

Data sharing is not applicable to this article as no new data were created or analyzed in this study. The source code for the proposed models will be available upon acceptance.

 \bibliographystyle{abbrvnat}

\bibliography{Reference}

\newpage 

\appendix

%
%
%

\section{Attention Mechanism}
\label{Attention_Section}
The attention mechanism was introduced to enable capturing long-range dependencies in sequential data.  Rather than compressing all past information into a single hidden representation, attention asks: \textit{Given a current query, which elements of the sequence are most relevant for the prediction?} This formulation allows the model to assign importance to different time steps based on content rather than position alone. The attention operation computes a weighted aggregation of input representations 
\begin{equation}
\text{Attention}(Q, K, V) = \text{softmax}\!\left(\frac{QK^{\top}}{\sqrt{d_k}}\right)V,
\label{eq:attention}
\end{equation}
where $Q$, $K$, and $V$ denote the query, key, and value matrices, respectively, and $d_k$ is the dimension of the key vectors. In this formulation, similarity between a query and all keys determines how much attention each input position receives. The scaling factor, $\sqrt{d_k}$, stabilizes gradients, and the softmax function converts similarity scores into normalized weights. Multiplying these weights by the value matrix produces a context vector, which is a weighted summary of the sequence that emphasizes the most informative elements. This mechanism enables the model to integrate information from both nearby and distant time steps in a flexible manner. 

Transformer architecture stacks multiple self-attention and feed-forward layers to model complex sequence relationships without relying on recurrence or convolution. Due to Transformers' lack of inherent notion of order, positional encodings are added to input embeddings to preserve temporal structure \cite{vaswani2017attention}. The encoder-decoder architecture further extends this framework by allowing the decoder to selectively attend to encoder outputs during prediction. This combination of flexibility, scalability, and parallelism has established Transformers as a foundation for state-of-the-art models across natural language processing, computer vision, and time-series forecasting \cite{lin2022survey, wen2022transformers}.

\section{Computational Complexity and Practical Adjustments of Transformers for Time-Series}

Self-attention requires quadratic computational and memory complexity in terms of the number of time steps in the input sequence. Thus, it could be computationally expensive for a long time-series. There exist many Transformer variants for time-series forecasting that use sparse, low-rank, or probabilistic attention to reduce computation and memory while still capturing important patterns in the data. Systematic surveys on Transformers provide comprehensive comparisons of different Transformer variants, highlighting their architectural trade-offs and relative performance across diverse forecasting tasks \cite{wen2022transformers, shan2025investigation}. \cite{zhou2025transformers} showed that attention models do not beat classical linear predictors. However, this claim might not hold for nonlinear scenarios such as the one considered in this study. Attention based model have shown promising results in similar forecasting problems \cite{baghaei2026transformer,peter2025forecasting, peter2025downscaling,el2025adaptive}.

\section{Genetic Algorithm Convergence Results}
\label{app:ga_convergence}

\begin{figure}
    \centering
    \includegraphics[width=1\linewidth]{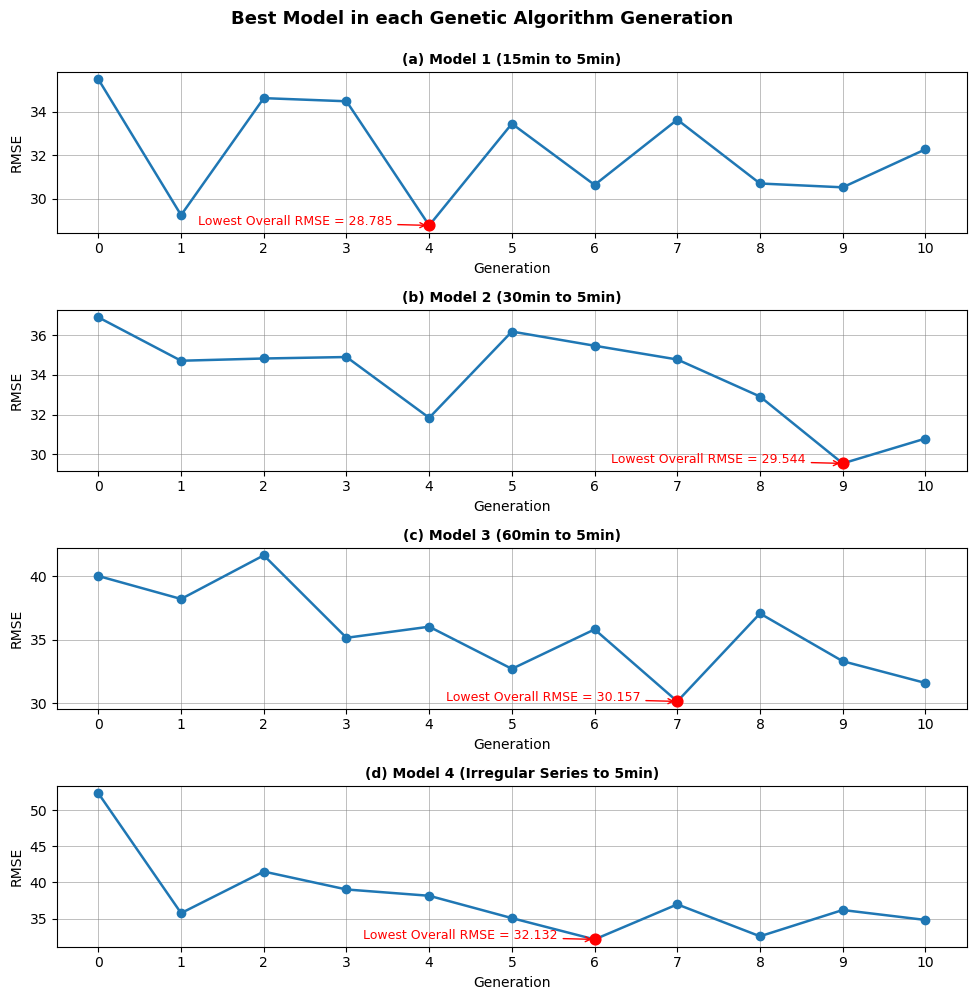}
    \caption{Best RMSE obtained in each genetic algorithm generation for different modeling scenarios. Generation 0 represents the initial population, and the red marker denotes the overall best-performing model in each case.}
\label{fig:ga_generation_rmse}
\end{figure}

Figure~\ref{fig:ga_generation_rmse} shows the evolution of the best RMSE obtained within each GA generation for the four input-resolution scenarios. Generation 0 represents the initial population, while subsequent generations correspond to models produced through the GA selection, crossover, and mutation process. For each generation, the lowest RMSE is reported to show how the best-performing candidate model changes across the optimization process.

The results indicate that the GA is able to identify improved model configurations over successive generations, although the improvement is not strictly monotonic. This non-monotonic behavior is expected because each generation contains a new set of candidate models, and the best model within a later generation is not always guaranteed to outperform the best model from an earlier generation. The best models are obtained in generation 4 for the 15-minute input model, generation 9 for the 30-minute input model, generation 7 for the 60-minute input model and, generation 6 for the irregular time series input model.

The figure shows that the GA search is able to identify improved configurations beyond the initial population, demonstrating its usefulness for optimizing the Transformer hyperparameters.

\section{Monthly Forecasting and Uncertainty Results}
\label{app:uncertainty_results}

\begin{figure}
    \centering
    \includegraphics[width=1\linewidth]{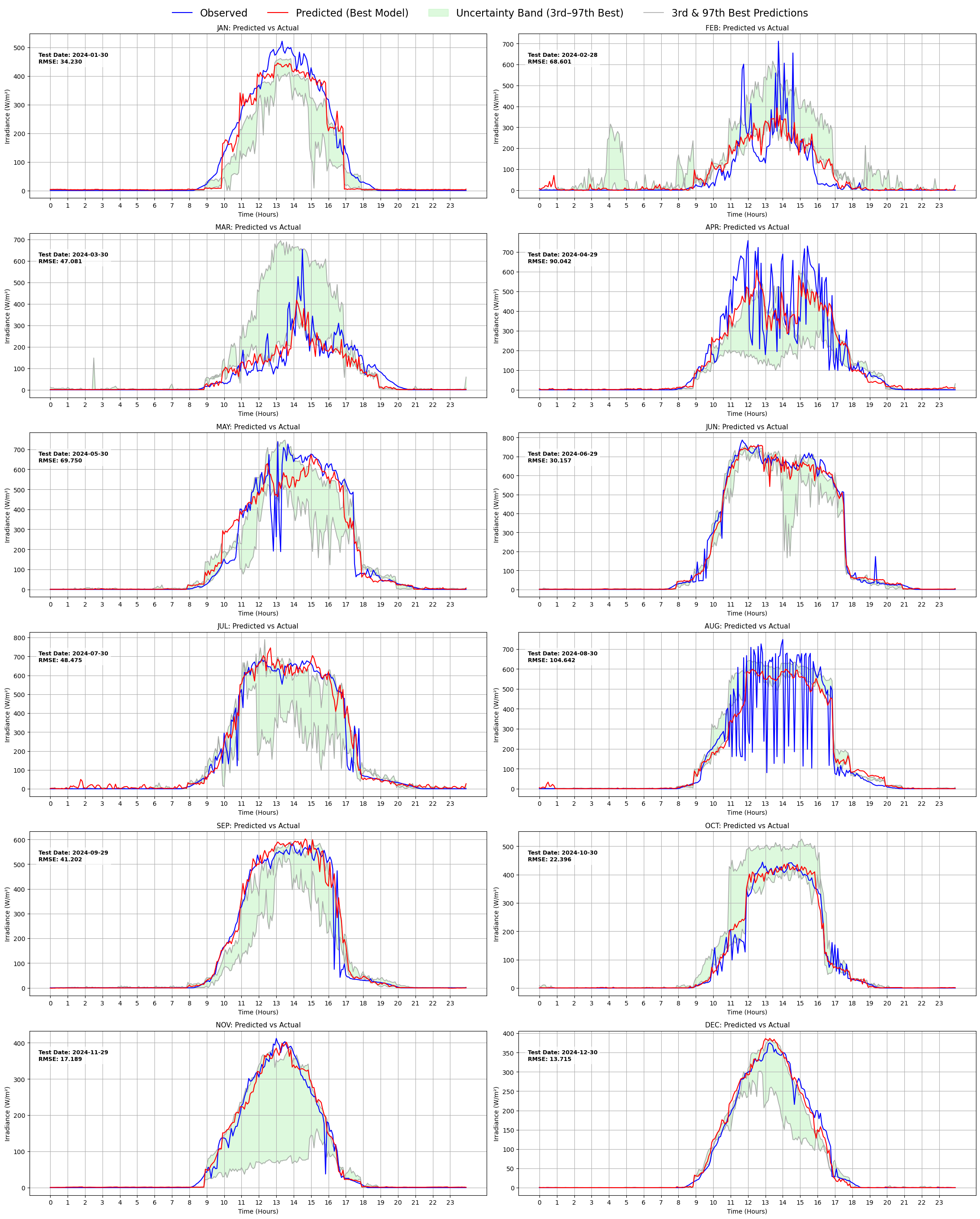}
    \caption{Monthly 5-minute local solar irradiance forecasts for 2024 using the proposed model. The observed values, best-model predictions, and uncertainty band between the 3rd-best and 97th-best retrained models are shown for each monthly test day.}
    \label{fig:all_months_plots_3rd_97th}
\end{figure}

\begin{figure}
    \centering
    \includegraphics[width=1\linewidth]{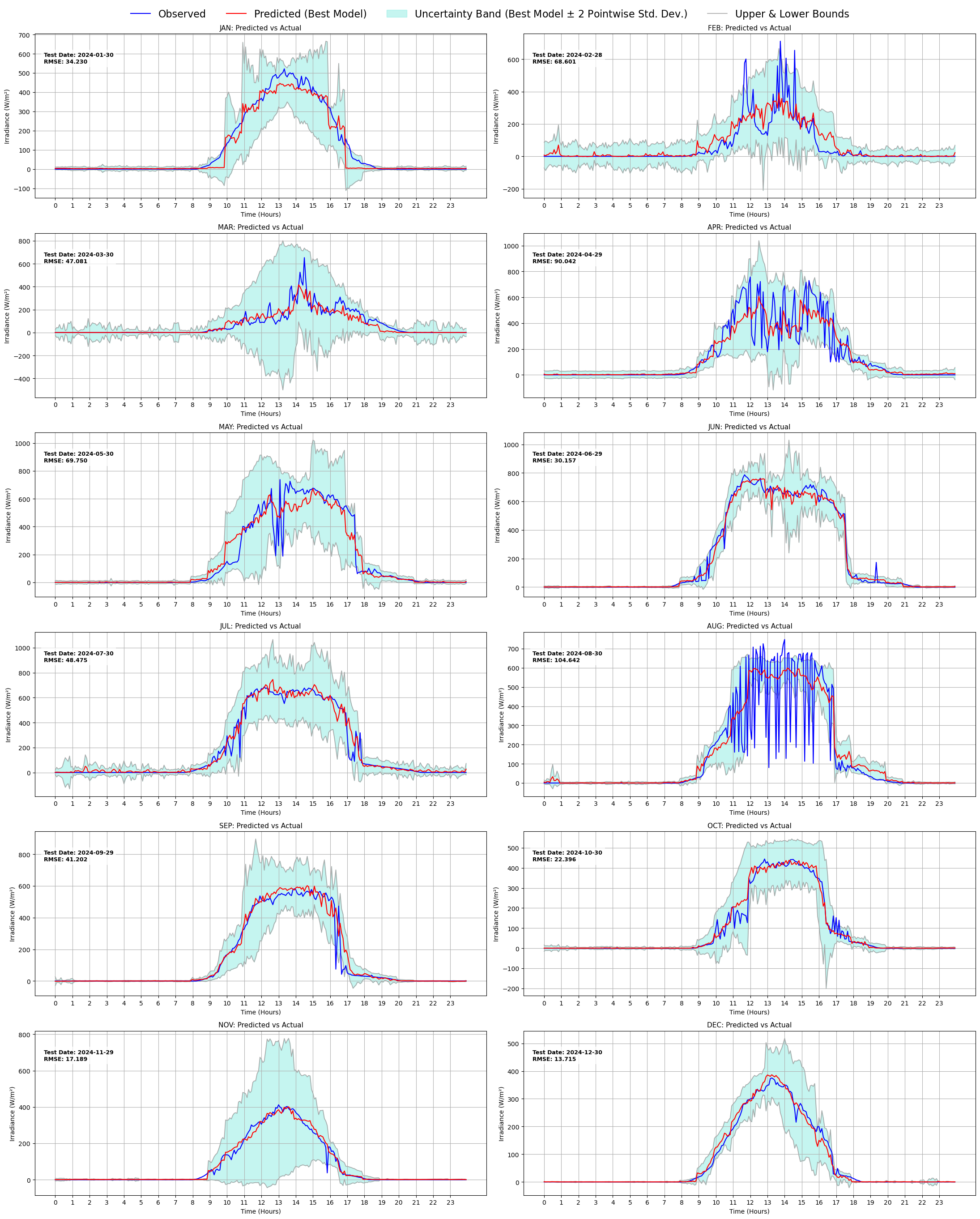}
    \caption
    n{Monthly 5-minute local solar irradiance forecasts for 2024 using the proposed model. The observed values, best-model predictions, and the pointwise uncertainty band around the best model are shown for each monthly test day.}
    \label{fig:all_months_pointwise}
\end{figure}

This section presents the monthly forecasting results for the year 2024 to further evaluate the performance of the proposed model across different seasonal conditions. For each month, the model was trained using the two months of historical data preceding the test date and was then used to predict the selected test day at a 5-minute resolution. The only exception is January, where only January data were available for training. The results for the predictions along with their uncertainty bounds are shown in Figure~\ref{fig:all_months_plots_3rd_97th} and~\ref{fig:all_months_pointwise}. 

All missing values in the dataset were filled using PCHIP interpolation, as described in Section~\ref{sec:data}. For November, five consistent periods of missing values were observed. In this case, the nighttime and early morning missing values were first set to zero because solar irradiance is expected to be zero during those periods, and the remaining missing values were then interpolated.

The monthly results show that the model is generally able to reproduce the overall daily pattern of local solar irradiance across the year. In most months, the predicted curves follow the observed irradiance reasonably well. However, the model has more difficulty capturing severe and abrupt spikes and dips in irradiance, particularly in months such as February, April, May, and August. These sharp fluctuations are likely caused by rapid cloud movement or short-term atmospheric changes. As a result, the predicted curves are often smoother than the observed curves and may underestimate sudden peaks or fail to fully capture rapid drops in irradiance. 

The uncertainty bands provide additional insight into the stability of the predictions across repeated retraining runs. Figure~\ref{fig:all_months_plots_3rd_97th} represents the range between the 3rd-best and 97th-best models from the 100 retraining runs, while Figure~\ref{fig:all_months_pointwise} represents the pointwise uncertainty band based on variance around the best model. Wider uncertainty bands are generally observed during daytime and peak irradiance periods, where the irradiance values are higher and more variable. In contrast, the uncertainty is usually narrower during nighttime, when irradiance is expected to remain close to zero.

These monthly results show that the proposed model can capture the broad seasonal and diurnal behavior of solar irradiance, but its performance decreases when the observed irradiance contains sudden, high-frequency fluctuations. This suggests that future improvements may require additional meteorological or cloud-related predictors to better capture abrupt irradiance spikes and drops.

\end{document}